\documentclass[sigconf]{acmart}
\renewcommand\footnotetextcopyrightpermission[1]{} 

\usepackage{bm}
\usepackage{algorithm}  
\usepackage[noend]{algpseudocode}
\usepackage{tabularx}
\usepackage{multirow}
\usepackage{subfigure}
\usepackage{footnote}\makesavenoteenv{tabular}\makesavenoteenv{table}
\usepackage{booktabs} 
\usepackage{threeparttable}

\copyrightyear{2018} 
\acmYear{2018} 
\setcopyright{acmlicensed}
\acmConference[MM '18]{2018 ACM Multimedia Conference}{October 22--26, 2018}{Seoul, Republic of Korea}
\acmBooktitle{2018 ACM Multimedia Conference (MM '18), October 22--26, 2018, Seoul, Republic of Korea}
\acmPrice{15.00}
\acmDOI{10.1145/3240508.XXXXXXX}
\acmISBN{978-1-4503-XXXX-XXXXXX}

\begin{document}
\title{Step-by-step Erasion, One-by-one Collection:\\A Weakly Supervised Temporal Action Detector}

\author{Jia-Xing Zhong, Nannan Li, Weijie Kong, Tao Zhang, Thomas H. Li, Ge Li}
\affiliation{%
  \institution{School of Electronic and Computer Engineering, Peking University}
}
\email{jxzhong@pku.edu.cn, linn@pkusz.edu.cn, weijie.kong@pku.edu.cn,} \email{t_zhang@pku.edu.cn, 18662197462@qq.com, geli@ece.pku.edu.cn}
\authornote{Corresponding author: Ge Li. This work was supported in part by the Project of National Engineering Laboratory-Shenzhen Division for Video Technology, in part by Science and Technology Planning Project of Guangdong Province, China (No. 2014B090910001), in part by the National Natural Science Foundation of China and Guangdong Province Scientific Research on Big Data (No. U1611461), in part by Shenzhen Municipal Science and Technology Program under Grant JCYJ20170818141146428, and in part by National Natural Science Foundation of China (No. 61602014). In addition, we would like to thank Jerry for English language editing.}

\begin{abstract}
Weakly supervised temporal action detection is a Herculean task in understanding untrimmed videos, since no supervisory signal except the video-level category label is available on training data. Under the supervision of category labels, weakly supervised detectors are usually built upon classifiers. However, there is an inherent contradiction between classifier and detector; \textit{i.e.}, a classifier in pursuit of high classification performance prefers top-level discriminative video clips that are extremely fragmentary, whereas a detector is obliged to discover the whole action instance without missing any relevant snippet. To reconcile this contradiction, we train a detector by driving a series of classifiers to find new actionness clips progressively, via \emph{step-by-step} erasion from a complete video. During the test phase, all we need to do is to collect detection results from the \emph{one-by-one} trained classifiers at various erasing steps. To assist in the collection process, a fully connected conditional random field is established to refine the temporal localization outputs. We evaluate our approach on two prevailing datasets, \textit{THUMOS'14} and \textit{ActivityNet}. The experiments show that our detector advances state-of-the-art weakly supervised temporal action detection results, and even compares with quite a few strongly supervised methods.
\end{abstract}

%
%



\maketitle

\section{Introduction}

\begin{figure}[!ht]
\centering
\includegraphics[width=0.45\textwidth]{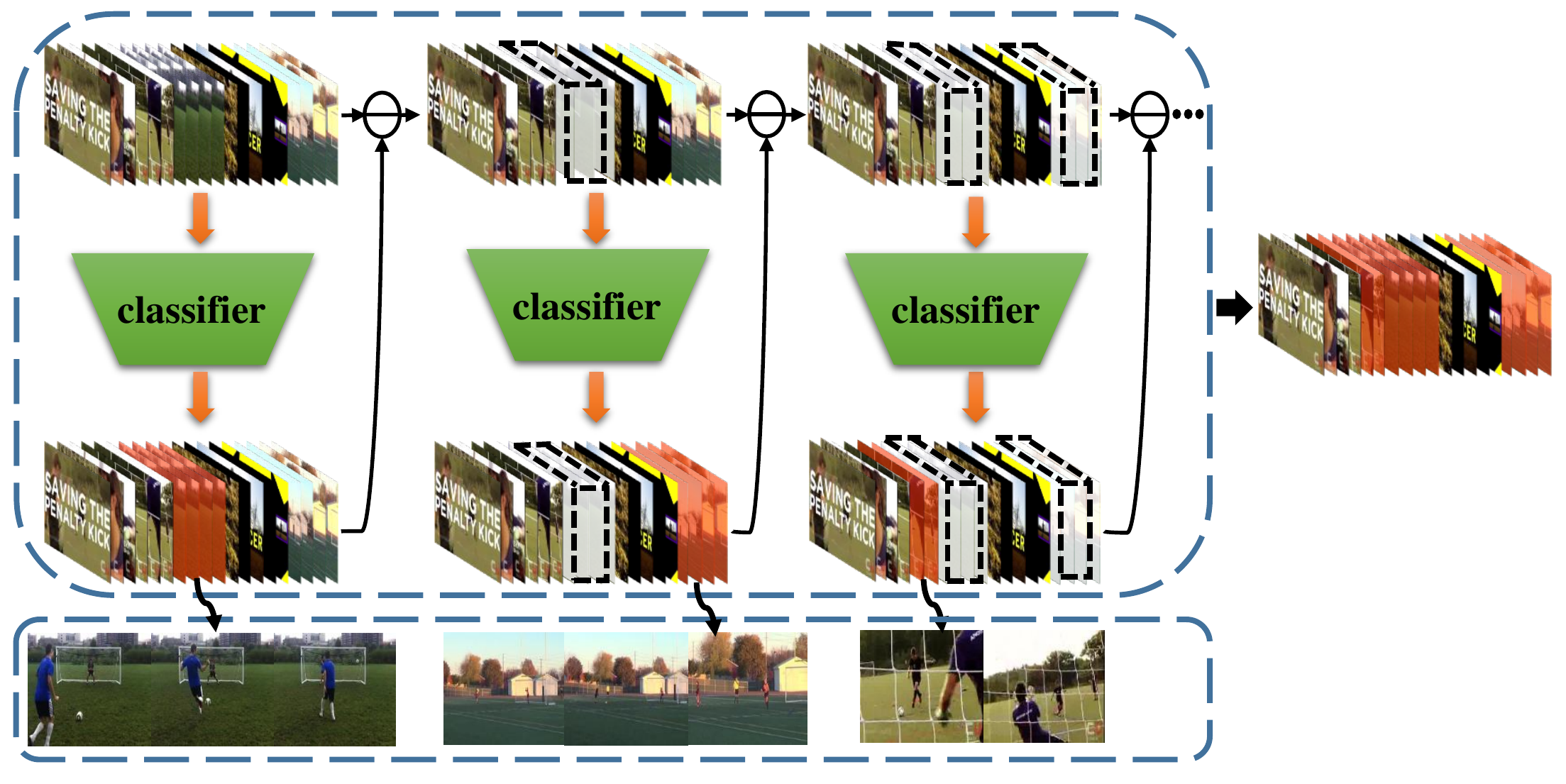}
\caption{\emph{Illustration of our detector.} A classification network firstly discovers the most discriminative video segments in response to the action ``shooting''. Then these mined snippets, marked in red, are erased from the training video. In this figure, the erased segments are marked in white and bordered with dotted lines at the next step. Another action classifier is trained on the remaining clips, which forces the classifier to explore other discernible snippets neglected by the previous one. We perform such processes for several rounds, and collect all mined video clips as the final temporal detection result.}
\label{example}
\end{figure}

During the past few years, action analysis has drawn much attention in the area of video understanding. There is an amount of research on this issue, based upon either hand-crafted feature representations~\cite{laptev2003space,wang2013action,wang2016mofap}, or deep learning model architectures~\cite{simonyan2014two,tran2015learning,longterm2015,karpathy2014}. A great deal of existing work handles action analysis tasks in a \textbf{\textit{strongly supervised}} manner, where the training data of action instance without backgrounds is manually annotated or trimmed out. In recent years, several strongly supervised methods have achieved satisfactory results~\cite{tran2015learning,DBLP:journals/corr/WangXWQLTG17,saha2016deep}. However, it is laborious and time-consuming to annotate precise temporal locations of action instances on increasingly large scale video datasets today. Additionally, as pointed out in~\cite{satkin2010modeling}, unlike object boundary, the definition of exact temporal extent of the action is often subjective and not consistent across different observers, which may result in additional bias and error. To overcome these limitations, utilizing the weakly supervised approach is a reasonable choice.

In this paper, we attempt to address the \textbf{\emph{temporal action detection}} problem, on which our model predicts the action category as well as the temporal location of action instance within a video. In the task of \textbf{\emph{weakly supervised learning}}, only video-level category label is provided as supervisory signal, and video clips containing action instances intermixed with backgrounds are untrimmed during the training process.
 
Detectors under weak supervision are often based on classifiers, since explicit labels are only available for classification of entire videos. However, a classifier differs strikingly from a detector. For purpose of better classification performance, a classifier desires to discover the most discriminative snippets that contribute most towards category correctness. Generally speaking, these top-level discriminative video clips are of short duration and temporally scattered. In contradiction to the classifier, a detector is supposed to find all video frames containing the certain action instance and hates any omission of ground truth. The contradiction between detector and classifier makes it difficult to fit a classification model to a detection task.

We deal with this contradiction by \textit{step by step} erasing clips with high classification confidence for several times in training. As illustrated in Figure~\ref{example}, the most discernible snippets about the action ``shooting'', such as ``penalty shots'', are likely to be removed at the first erasion step. In this case, the classifiers trained at subsequent steps have no choice but to seek other relevant clips such as ``midfielder's shots'' or ``scoring goals'', \textbf{\textit{since the top-level discriminative video segments have been deleted and are invisible to these classifiers}}. By erasing discernible clips \textit{step by step}, classifiers trained at different steps are capable of finding different actionness snippets. In the test phase, we only need to collect detection results from the \emph{one-by-one} classifiers at various erasing steps. Consequently, the fusion of erased video snippets during the whole detection process constitutes the integral temporal duration of an action. However, limited by the representative ability of classifiers, our model might misclassify a handful of clips. To assist in collecting detection results from the \textit{one-by-one} classifiers, we further establish a fully connected conditional random field (FC-CRF)~\cite{crf2011}, in order to retrieve the ignored actionness snippets as well as mitigate detection noises. Particularly, our FC-CRF endows the detector with the prior knowledge that the extent of action instance on a temporal domain should be continuous and smooth. Based on this prior knowledge, the FC-CRF is helpful in connecting separated actionness clips and deleting isolated false-positive detection results.

In a nutshell, our main contributions in this paper are as follows:
\begin{itemize}
\item We present a weakly supervised model to detect temporal action in untrimmed videos. The model is trained  with \emph{step-by-step} erasion on videos to obtain a series of classifiers. In the test phase, it is convenient to apply our model by collecting detection results from the \emph{one-by-one} classifiers.
\item To our best knowledge, this is the first work that introduces the FC-CRF to temporal action detection tasks, which is utilized to combine the prior knowledge of human beings and vanilla outputs of neural networks. Experimental results show that the FC-CRF boosts detection performance by \(20.8\%\) mAP@0.5 on \textit{ActivityNet}.
\item We carry out extensive experiments on two challenging untrimmed video datasets, \textit{i.e.}, \textit{ActivityNet}~\cite{activitynet} and \textit{THUMOS'14}~\cite{thumos14}; the results show that our detector achieves comparable performance on temporal action detection with many strongly supervised approaches.
\end{itemize}

\begin{figure*}[!t]
\centering\includegraphics[width=0.8\textwidth]{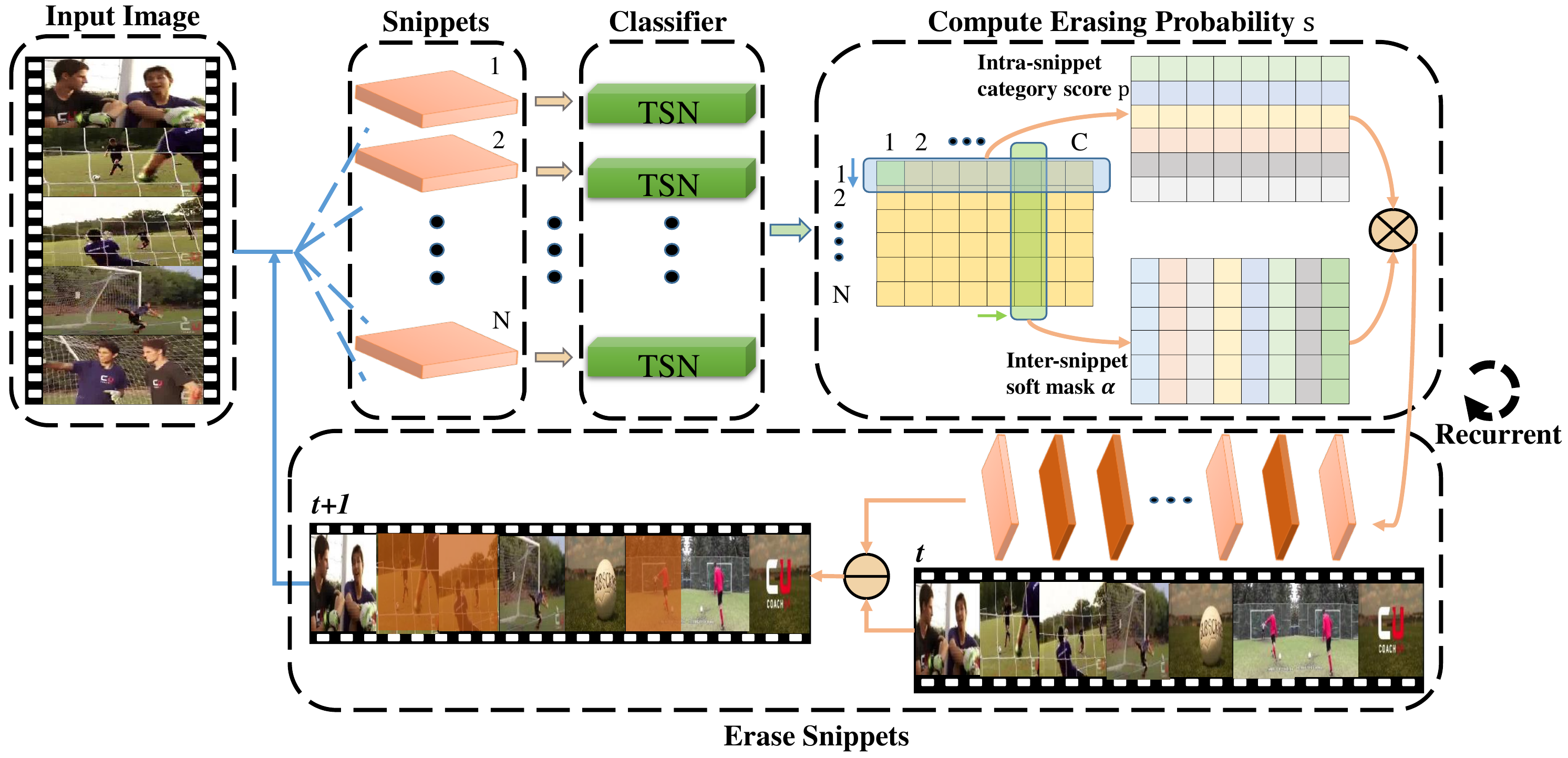}
\caption{\textit{Overview of training process with step-by-step erasion.} The input video is averagely divided into non-overlapping snippets, and fed into a classifier (\textit{e.g.}, TSN) to obtain snippet-wise responsive scores. Based on the scores, we compute the erasing odds for every snippet by applying a soft mask to category probability. Afterwards these snippets are removed with their erasing probabilities. At the next step, another classifier is trained on the remaining video data according to such a strategy, and it is expected to discover other actionness snippets missed by the previous classifier. We repeat the cycle for several times until no useful clips are revealed.}
\label{pipleline}
\end{figure*}

\section{Related Work}
\textbf{Action Recognition \& Temporal Detection with Deep Learning.}
During the past few years, driven by the great success of deep learning in the computer vision area~\cite{Zhong2017,xudan2018tpami,Rao_2017_ICCV}, a number of models~\cite{karpathy2014,wang2015trajectory,twostream2016cvpr,tran2015learning,simonyan2014two,qiu2017learning,Tang_2018_CVPR} with deep architectures, especially Convolutional Neural Network (CNN) or Recurrent Neural Network (RNN), have been introduced to video-based action analysis. Karpathy \textit{et al.}~\cite{karpathy2014} first employ deep learning for action recognition in video, and design a variety of deep models which process a single frame or a sequence of frames. Tran \textit{et al.}~\cite{tran2015learning} construct a C3D model, which executes 3D convolution in spatial-temporal video volume and integrates appearance and motion cues for better representation. Wang \textit{et al.}~\cite{DBLP:journals/corr/WangXWQLTG17} propose Temporal Segment Network (TSN), which inherits the advantage of the two-stream feature-extraction structure, and leverage sparse sampling scheme to cope with longer video clip. Qiu \textit{et al.}~\cite{qiu2017learning} present pseudo-3D (P3D) residual networks to  recycle off-the-shelf 2D networks for a 3D CNN. Carreira and Zisserman considerably improve performance in action recognition by pretraining Inflated 3D CNNs (I3D) on \textit{Kinetics}. Apart from dealing with action recognition, there are some other work to address action detection or proposal generation \cite{DBLP:journals/corr/HuangLZL17,shou2016temporal,Shou2017CDC,zhu2016efficient,xiong2017pursuit,gao2017turn,yeung2016end,li2016proposal,Carreira2017Quo,Zhao2017Temporal,Lin2017Single,Lea2017Temporal,Heilbron2017SCC,Xu2017R,Yuan2017Temporal,Dai2017Temporal}. Shou \textit{et al.}~\cite{shou2016temporal} utilize a multi-stage CNN detection network for temporal action localization. Escorcia \textit{et al.}~ \cite{daps2016} propose DAPs model which encodes the video sequence with RNN and retrieves action proposals at a single process. Lin \textit{et al.}~\cite{Lin2017Single} skip the proposal generation step with a single shot action detector (SSAD). Shou \textit{et al.}~\cite{Shou2017CDC} devise the Convolutional-De-Convolutional (CDC) network to determine precise temporal boundaries. Our approach differs from the aforementioned works: they build deep learning models upon precise temporal annotations or trimmed videos, whereas our model directly employs the untrimmed video data for training and requires only video-level category labels.

\textbf{Weakly Supervised Learning in Video Analysis.}
Although strongly supervised methods make up the bulk of the solutions to video analysis tasks, there is some research work \cite{laptev2008movie,weakly2014eccv,modeling2016eccv,webly2016eccv,Kuehne:2017:WSL:3170020.3170285,alignment2015iccv,wang2018untrimmed,Sun2015Temporal} which adopt weakly supervised approach to action analysis in video. The supervisory information used within those methods for conducting the training includes: movie scripts \cite{laptev2008movie,duchenne2009video}, temporally ordered action lists \cite{weakly2014eccv,modeling2016eccv}, video-level category label \cite{wang2018untrimmed} or web videos and images \cite{webly2016eccv}, \textit{etc}. Laptev \textit{et al.}~\cite{laptev2008movie} and Marszalek \textit{et al.}~\cite{action2009marszalek} focus on mining training samples from movie scripts for action recognition, without applying an accurate temporal alignment of the action and respective text passages. Huang \textit{et al.}~\cite{modeling2016eccv} address action labeling by introducing the extended connectionist temporal classification framework (CTC) adapted from language model to evaluate possible alignments. Sun \textit{et al.}~\cite{Sun2015Temporal} apply cross-domain transfer between video frames and web images for fine-grained action localization. Wang \textit{et al.}~\cite{wang2018untrimmed} establish the UntrimmedNets to work on weakly supervised action detection problem. The work proposed in \cite{hide2017seek} shares a similar training strategy with our approach, and the difference is that it trains a single classifier by randomly hiding video snippets for the \emph{localization} task, while we focus on the \emph{detection} task by recurrently training a series of classifiers.

Our approach draws the inspiration from the work proposed in \cite{erase2017}, which applies the erase-and-find strategy to image-based semantic segmentation in a weakly supervised manner. It recurrently trains a set of classifiers to discover the discriminative image regions related to a specific object. This inspired us to develop an erase-and-find method for video understanding. The core difference on two learning strategies is that it needs to additionally train a strongly supervised segmentation network using pixel-wise pseudo labels generated by these classifiers, whereas we directly collect the outputs from the series of trained classification networks for prediction. Our approach decentralizes the detection task to several disparate classification networks, so there is no requirement for our detector to train any extra strongly supervised model.

\section{Step-by-step Erasion, One-by-one Collection}

Our model consists of two parts: \textbf{\textit{training with step-by-step erasion on videos}} and \textbf{\textit{testing by collecting results from one-by-one classifiers}}. During the training process, we progressively erase the snippets with high confidence of action occurrence. By doing so, we obtain a series of classifiers with respective predilections for different types of actionness clips. In the test phase, we iteratively select snippets with action instances based on the trained classifiers, and refine the fused results via an FC-CRF.     

\subsection{Training with Step-by-step Erasion}\label{train_text}
As shown in Figure~\ref{pipleline}, we alternate with 3 operations: \textit{erasing probability computation}, \textit{snippet erasion} and \textit{classifier training} for several rounds. Suppose that a video \(V=\{v_n\}_{n=1}^{N}\) contains \(N\) clips, with \(K\) video-level category labels \(Y=\{y_k\}_{k=1}^{K}\). Given a snippet-wise classifier specified by parameters \(\theta\), we can obtain the vanilla classification score \(\phi(V;\theta)\in\mathbb{R}^{{N}\times{C}}\), where \(C\) is the number of all categories.

At the \(t^{th}\) erasing step, we denote the remaining clips of a training video as \(V^{t}\) and represent the classifier as \(\theta^{t}\). For the \(i^{th}\) row \(\phi_{i, :}\) of \(\phi(V^{t};\theta^{t})\), corresponding to the raw classification score of the \(i^{th}\) clip, we compute the intra-snippet probability of the \(j^{th}\) category with \emph{softmax} normalization:

\begin{equation}\label{eq:pij}
  p_{i,j}(V^t)=\frac{exp({\phi_{i, j}})}{\sum_{c=1}^{C}exp({\phi_{i, c}})} \,.
\end{equation}
\begin{algorithm}[h]  
  \caption{Training with Step-by-step Erasion}  
  \label{alg:1}  
  \begin{algorithmic}[1]  
    \Require
      \(\theta^{0}\): initial snippet-wise classifier;
      \(\tau\): discounting threshold about the soft mask; 
      \(D^{0}=\{(V^{0}, Y) \mid V^{0}=\{v_n\}_{n=1}^{N}, Y=\{y_k\}_{k=1}^{K}\}\): training set
    \Ensure  
      \(\{\theta^{t}\}_{t=1}^T\): trained models at various steps   
    \State 
      Initialize sequence number of erasing step \(t=1\) and trained model count \(T=0\)
    \Repeat  
      \State Train the classifier \(\theta^{t}\) from \(\theta^{t-1}\) with \(D^{t-1}\)
      \State Initial \(D^{t}=\emptyset\)
      \For {each video \(V^{t-1}\) in \(D^{t-1}\)}
        \State Initial \(V^{t}=V^{t-1}\)
        \State Compute the classification score \(\phi(V^{t}; \theta^{t})\) 
        \For {\(y_j \in Y\)}
             \For {\(v_i \in V^{t}\)}
                \State Compute \(s_{i,j}(V^t)\) as Eq.~\eqref{eq:sij}
                \State Generate a sequence \(\epsilon\) of \(N\) random values within [0 ,1]
                \State Obtain erasing clips: \(E=\{v_{i} \mid s_{i,j}(V^t)>{{\epsilon}_i}\}\)
                \State Erase clips from the video: \(V^{t}=V^{t} \setminus E\)
              \EndFor
        \EndFor
        \State Update training data: \(D^{t}=D^{t} \cup (V^{t}, Y)\)
      \EndFor
      \State Update states: \(T=T+1\); \(t=t+1\)
    \Until {no useful clips are found}  
  \end{algorithmic}  
\end{algorithm}

In practice, the \emph{softmax} transformation may amplify noisy activation responses for background clips. Moreover, solely modeling a single snippet is not enough to harness global information among different clips in the whole video. To amend the intra-snippet probability, we present an inter-snippet soft mask mechanism. For the \(j^{th}\) column \(\phi_{:, j}\) representing the confidence of the \(j^{th}\) category over all clips, we apply \emph{min-max} normalization to them. Although a background clip may have its own highest activation response to one certain category, \textbf{\textit{the responsive intensity is likely lower than its ground-truth peers with such kind of action instance}}. The \textit{min-max} operation substantially suppresses the score of background clips whose category responses are relatively weak. Therefore, we define the inter-snippet soft mask w.r.t. the \(j^{th}\) category upon the \(i^{th}\) clip as:
\begin{equation}
  \alpha_{i,j}(V^t)=\delta_\tau(\frac{\phi_{i, j}-\min\phi_{:, j}}{\max\phi_{:, j}-\min\phi_{:, j}}) \,,
\end{equation}
where \(\delta_\tau\) rescales the result of the \textit{min-max} normalization upon a discounting threshold \(\tau\in(0,1]\):
\begin{equation}
  \delta_\tau(\bm\cdot)=
  \begin{cases}
    1 & \text{if } {\bm\cdot} > {\tau} \,;\\
    \frac{\bm\cdot}{\tau} & \text{otherwise} \,. \\
  \end{cases}
\end{equation}
The discounting threshold \(\tau\) determines how much rigorous the erasing standard we formulate: the larger \(\tau\) implies the less video clips are removed. Hence, \(\alpha_{i,j} \in [0,1]\) constitutes a soft mask. Unlike many attention mechanisms learned from neuron parameters, this inter-snippet mask needs no extra surgery on neural networks, and it can mitigate the noise from background clips in a simple way. Finally, we compute the erasing odds by element-wise multiplying the category probability with the soft mask:
\begin{equation}\label{eq:sij}
  s_{i,j}(V^t)=\alpha_{i,j}(V^t)p_{i,j}(V^t) \,.
\end{equation}

By the end of current erasing step \(t\), we remove snippets according to their erasing probability \(s\) from the remaining video, and utilize the rest snippets to train a new classifier at the next erasing step \(t + 1\). During the whole training process, we repeat such erasing steps to gradually find out discriminative snippets as in \textbf{Algorithm~\ref{alg:1}}.

Ideally, we would stop the training process when no more useful video clips can be discovered. However, it is impossible to make such a perfect decision in reality, because using only the video-level category labels is insufficient to provide temporal information. In preliminary experiments, we have found that the \textit{excessive erasion} introduces a spate of fragmentary snippets that are helps little in making up an integral segment with action instance. In other words, \textbf{\emph{scattered video clips mined with excessive erasion are hardly combined into a continuous segment}}. Hence, the normalized number of integral erased segments with the \(j^{th}\) category at the \(T^{th}\) step is a useful criterion: 

\begin{equation}\label{eq5}
m_j^T=\frac{|M_j^T|}{|M_j^1|}\,,
\end{equation}
where \(M_j^T\) is composed of video segments with continuous clips removed up to the \(T^{th}\) step, and its cardinality is normalized by \(|M_j^1|\) to alleviate the interference of various action durations. At the \(T^{th}\) step, we stop erasing for the \(j^{th}\) class if \(m_j^T\) nearly no longer changes, and reserve classifiers up to the \((T-1)^{th}\) step. Although the terminal criterion \(m_j^T\) is just based on our empirical observation, it is effective in practice, which we will elaborate in the Section~\ref{experiments}.
\subsection{Testing with One-by-one Collection}\label{text_test}
\begin{figure}[!t]
\centering\includegraphics[width=0.43\textwidth]{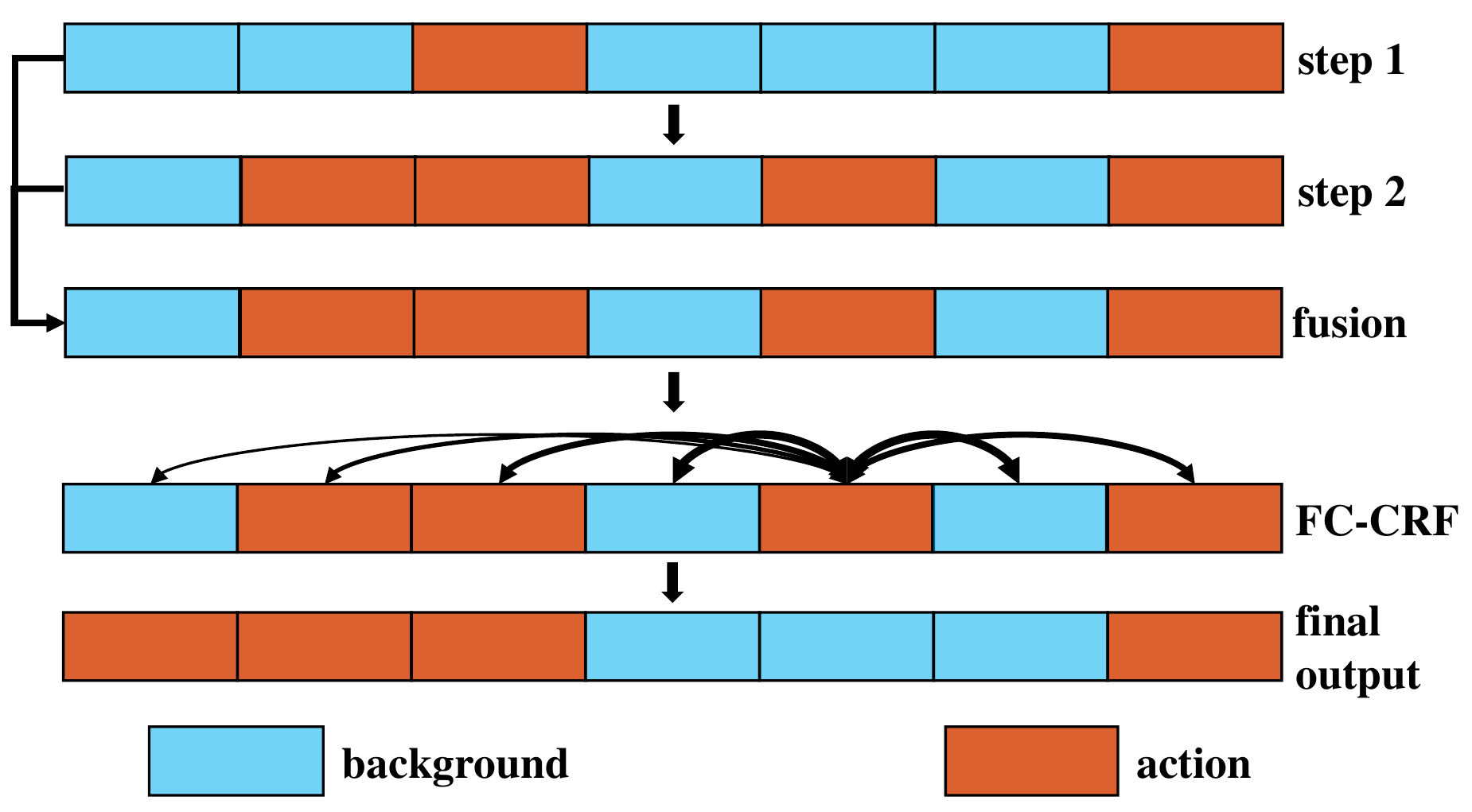}
\caption{\textit{Testing with one-by-one collection.} First of all, we iteratively collect predicted clips from \textit{one-by-one} trained classifiers. Then average fusion is adopted over the results. Finally, the category probability of all clips are refined by an FC-CRF to incorporate prior knowledge and classifier outputs.}
\label{CRF}
\end{figure}

As Figure~\ref{CRF} depicts, we collect the results from the \textit{one-by-one} trained classifier, and refine them with an FC-CRF. In the test phase, we have obtained several trained classifiers \(\{\theta^{t}\}_{t=1}^T\) from \textbf{Algorithm~\ref{alg:1}}. Our basic idea is to iteratively fetch snippets with high erasing score from \textit{one-by-one} classifiers, and fuse them together as the final detection results.

Denote a video \(V\) as a sequence of \(N\) clips \(\{v_n\}_{n=1}^{N}\). It is natural to take the average of the category probability \(p\) and the soft mask value \(\alpha\) over the T steps for the \(i^{th}\) clip of the \(j^{th}\) category as:

\begin{equation}\label{eq6}
\overline{\alpha}_{i,j}(V)=\frac{1}{T}\sum_{t=1}^{T}{\alpha_{i,j}}(V)\,,
\end{equation}
\begin{equation}\label{eq7}
\overline{p}_{i,j}(V)=softmax(\sum_{t=1}^T\log{p_{i,j}(V))}\,,
\end{equation}
where variable definitions on the \textit{right-hand side} of equations follow the subsection~\ref{train_text}, and the detection confidence \(\bar{s}=\bar{p}\bar{\alpha}\) can be readily computed.

However, the representative ability of video-based classifiers is still imperfect nowadays. Accumulated misclassified results over \textit{one-by-one} classification networks will severely degrade the detection performance. Thus, the direct collection of outputs from these multi-step classifiers is powerless to delineate the complete and precise temporal location. Due to this limitation of the classifiers, it is imperative to refine the average results through our prior knowledge.

As pointed out in~\citep{Mobahi2009Deep,Wiskott2002Slow,Jayaraman2016Slow}, the temporal coherence is ubiquitous in videos. In other words, the temporally vicinal video clips tend to contain similar information, and the actionness extent on time domain should be continuous. Therefore, \textbf{\emph{neighbor snippets are inclined to have the same label}}. We would like to impart this knowledge to an FC-CRF~\cite{fc_crf}. To our best knowledge, the FC-CRF is first introduced to video-based temporal action detection in this paper. In the formulation of conventional linear-chain CRFs, only the relationship between adjacent nodes is modeled. Unlike linear-chain CRFs, our FC-CRF takes into consideration the relationship between any and all nodes, in order to make full use of global information in a video. On the whole, our FC-CRF employs the Gibbs energy function of a label assignment \(l=\{l_1, l_2, ... , l_{N-1}, l_N\}\) as:
\begin{equation}
  E(l)=\sum_{i=1}^N \psi_{u}(l_i)+\sum_{i \neq j}^N \psi_{p}(l_i, l_j)\,,
\end{equation}
where \(l_i\) and \(l_j\) are category labels of the \(i^{th}\) and \(j^{th}\) clips. The two terms on \textit{right-hand side} respectively represent classifier predictions and prior knowledge. We compute the first term upon unary potential \(\psi_{u}(l_i)=-\log \overline{p}_i\), where \(\overline{p}_i=\{\overline{p}_{i,1}, \overline{p}_{i,2}, ..., \overline{p}_{i,C-1}, \overline{p}_{i,C}\}\) is the \(i^{th}\) component of average classification probability \( \overline{p}\) obtained from Eq.~\eqref{eq7}. The second term is based on pairwise potential \(\psi_{p}(l_i, l_j)\) between arbitrary clip pairs \(i\) and \(j\), expressed as:
\begin{equation}
  {\psi_{p}(l_i, l_j)=\omega\mu(l_i, l_j)exp({-\frac{{\Vert i-j \Vert}^2}{2\sigma^2}})}\,,
\end{equation}
where the compatibility function is determined as in the Potts model, \textit{i.e.}, \(\mu(l_i, l_j)=1\) if \(l_i \neq l_j\), otherwise \(\mu(l_i, l_j)=0\). That is to say, we only penalize nodes in the FC-CRF with distinct labels. We encourage snippets \(i\) and \(j\) in temporal proximity to be assigned the same label, with a Gaussian kernel. Intuitively, our Gaussian kernel exerts an influence between any two snippets, and the influence has an exponential decay as the temporal distance increases. There are two hyper-parameters of the FC-CRF: \(\omega\) is the fusion weight to balance unary and pairwise potentials, and \(\sigma\) controls the scale of Gaussian kernel.

\begin{figure}[!t]
\centering\includegraphics[width=0.45\textwidth]{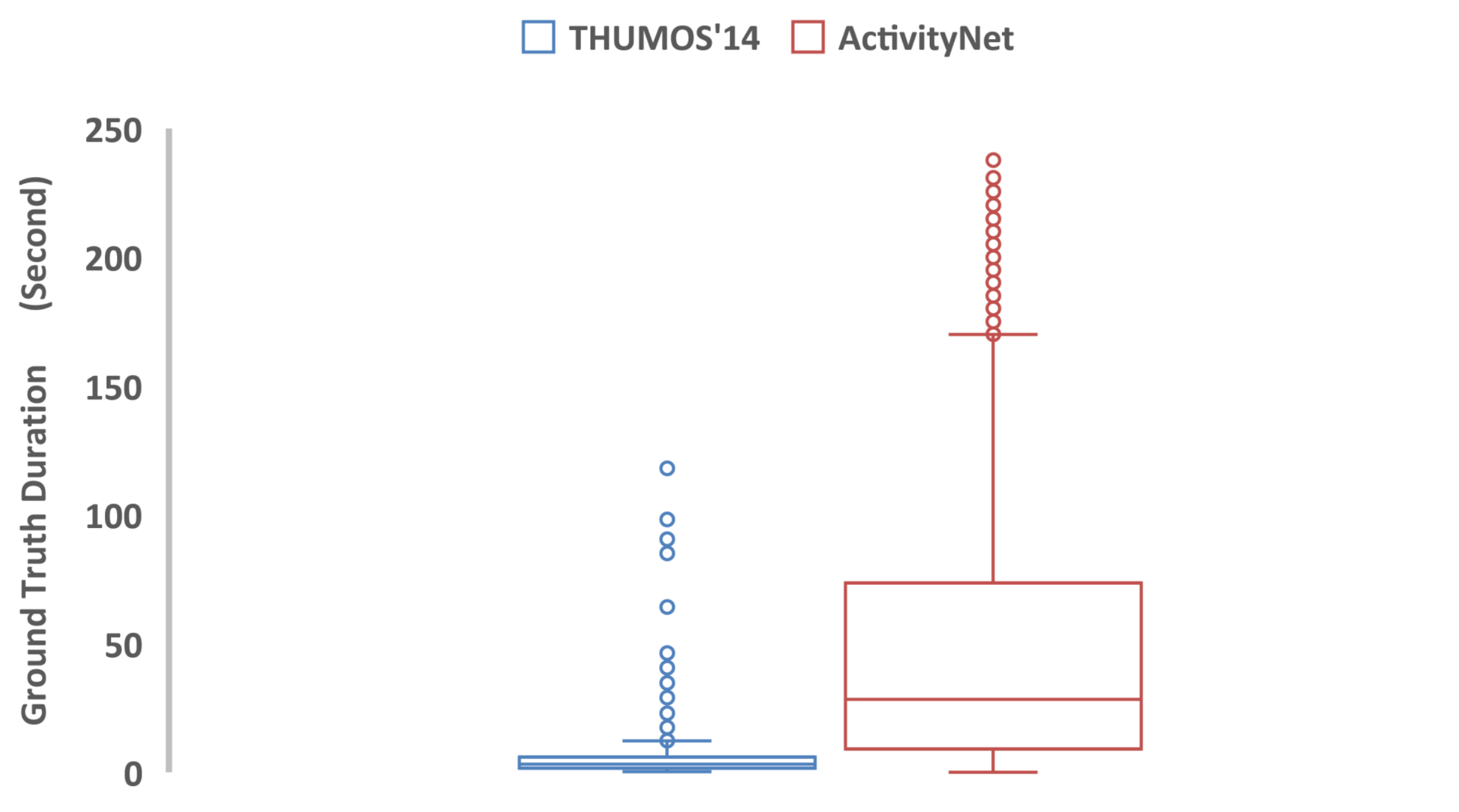}
\caption{\textit{Box-whisker plot of ground truth duration.} The time span of ground truth on \textit{THUMOS'14} is much shorter than that on \textit{ActivityNet}. In particular, the median duration of actions on \textit{THUMOS'14} is 3.1 seconds, while that on \textit{ActivityNet} is 28.3 seconds. Due to this fact, we adopt different sampling strategies and soft mask settings.}
\label{fig:boxplots}
\end{figure}

After establishing an FC-CRF with the Gibbs energy \(E(l)\), we approximate probabilistic inference with mean field as in \cite{fc_crf}, and compute the refined category probability \(\widetilde{p}_{i}\) for the \(i^{th}\) clip. According to this probability, we select the clips whose \(\widetilde{s}_{i,j}=\overline{\alpha}_{i,j}\widetilde{p}_{i,j}>0.5\) as final temporal detection results.
\section{Experiments}\label{experiments}
In this section, we first introduce the datasets and our implementation. Then we dive deeper in details of the proposed temporal action detector, including ablation studies, training terminal criterion and stability of hyper-parameters. Finally, we report our temporal detection results, and make comparisons to state-of-the-art approaches.

\subsection{Datasets}
We conduct our experiments on two prevailing datasets comprised of untrimmed videos, \textit{i.e.}, \textit{THUMOS'14}~\cite{thumos14} and \textit{ActivityNet}~\cite{activitynet}. Note that \textbf{\emph{we only use video-level category labels as our supervisory signal}} in training, albeit both datasets are annotated with the temporal action boundaries.

\textbf{\textit{THUMOS'14}} has 101 classes with 18,394 videos, a subset of which with 20 action categories is employed for temporal action detection tasks. Following~\cite{Zhao2017Temporal}, two falsely annotated videos (270, 1496) on
the test set are excluded in the experiment. In general, every video has a primary action category. Additionally, some videos may contain one or more action instances from other classes. Following the previous temporal detection work~\cite{gao2017turn,xiong2017pursuit,wang2018untrimmed}, we use the validation set for training, and evaluate our detection performance on the test set. 

\textbf{\textit{ActivityNet}} is a challenging benchmark for action recognition and temporal detection with a 5-level class hierarchy. We conduct experiments on its version 1.2, which has 100 classes with 9,682 videos, including 4,819 training videos, 2,383 validation videos, and 2,480 test videos. On \textit{ActivityNet}, each video belongs to one or more action categories as \textit{THUMOS'14}. Following works~\cite{xiong2017pursuit,Zhao2017Temporal} on \textit{ActivityNet v1.2}, we train our detector on the training data and test it on the validation set.

As for \textbf{evaluation metrics}, we follow the standard protocol, reporting mean Average Precision (mAP) at different temporal Intersection-over-Union (tIoU) thresholds. In such a formulation, the temporal action detection task can be viewed as an information retrieval problem. For every action category, all predicted video clips on the test set are ranked by detection confidence. The prediction for a certain class is deemed to be correct if and only if its tIoU with ground truth is greater than or equal to the threshold, and the mAP is defined upon these correct predictions. Both datasets have their own convention of tIoU thresholds since they originate from two competitions respectively. On the \textit{
THUMOS'14}, the tIoU thresholds are \(\{0.1, 0.2, 0.3, 0.4, 0.5\}\). On \textit{ActivityNet}, the tIoU thresholds are \(\{0.5, 0.75, 0.95\}\), and the average mAP at theses thresholds is also reported.\footnote{Strictly speaking, the average mAP is practically calculated with tIOU thresholds \([0.5 : 0.05 :
0.95]\).} 

\subsection{Implementation Details}
We implement our algorithm on the \textit{Caffe}~\cite{caffe}, and choose \textit{TSN}~\cite{DBLP:journals/corr/WangXWQLTG17} as our backbone classification network. For sake of an apples-to-apples comparison, we keep identical settings with \textit{UntrimmedNet}~\cite{wang2018untrimmed}: \(batch\_size=256\), \(momentum=0.9\), \(weight\_decay=0.0005\), and normalize the label with \(\ell 1\)-norm~\cite{l1_norm} for multi-label videos. Before erasion, we initially train our model for decent classification performance. In the step-by-step erasion phase, the max iteration number is 8,000 for both streams of \textit{TSN} at each erasing step, and we stop training as soon as the classification network converges on the validation data. We repeat erasing processes for \(4\) times at most on the two datasets. During the training process, the base learning rate is 0.0001 for the spatial stream, and decreases to one-tenth of the original learning rate every 1,500 iterations. For the temporal stream, we set the base learning rate as 0.0002, with the same decay strategy as the spatial stream.

As shown in Figure~\ref{fig:boxplots}, the duration of ground truth on \textit{Thumos'14} is evidently shorter than that on \textit{ActivityNet}. To this end, the experiment settings are slightly different between these two datasets. For \textit{ActivityNet}, we sample snippet scores every 15 frames as a detection snippet and apply soft mask threshold \(\tau=0.001\) to keep more snippets. In the case of \textit{THUMOS'14}, we extract detection snippets at intervals of 5 frames and use a more rigorous mask threshold \(\tau=0.5\), since its ground truths last a shorter time.

\subsection{Experimental Verification \& Investigation}
In this subsection, we investigate the further details of the presented model in three respects. For training, we firstly focus on the necessity of \emph{soft mask} and the significance of \emph{step-by-step erasion}. In addition, the \emph{criterion for training termination} is evaluated. For testing, we explore the stability of \emph{hyper-parameters} in FC-CRF and verify the effectiveness of \emph{FC-CRF} in collection procedure.

\begin{figure}[!t]
\centering
\subfigure[\textit{THUMOS'14}.] {
\includegraphics[width=0.33\textwidth]{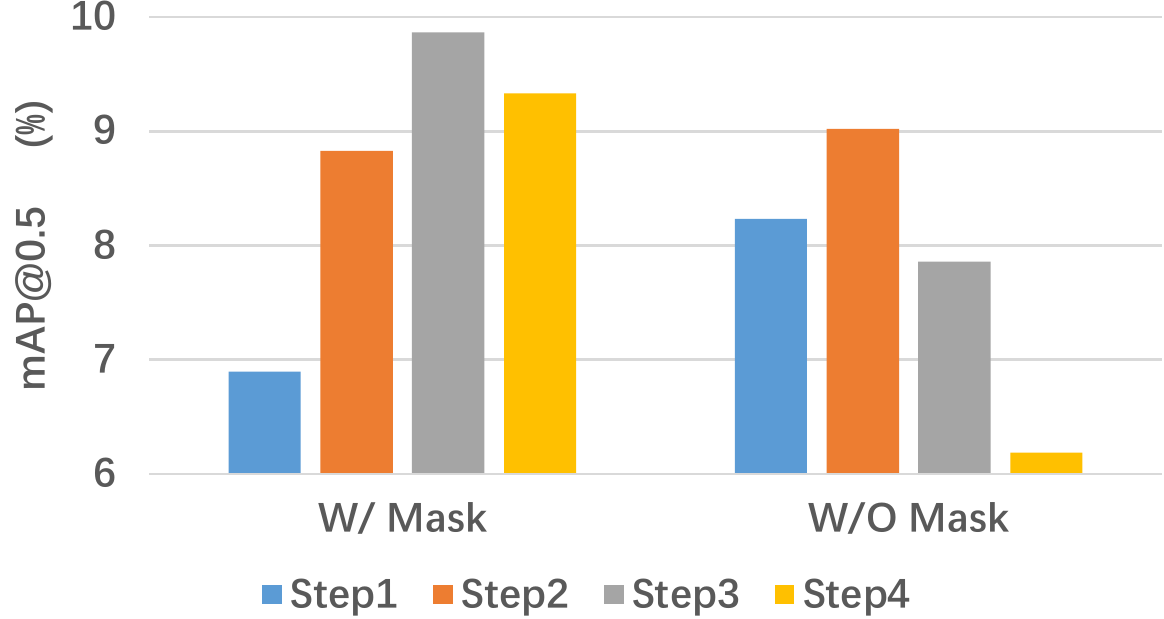} 
} 
\subfigure[\textit{ActivityNet}.] {
\includegraphics[width=0.33\textwidth]{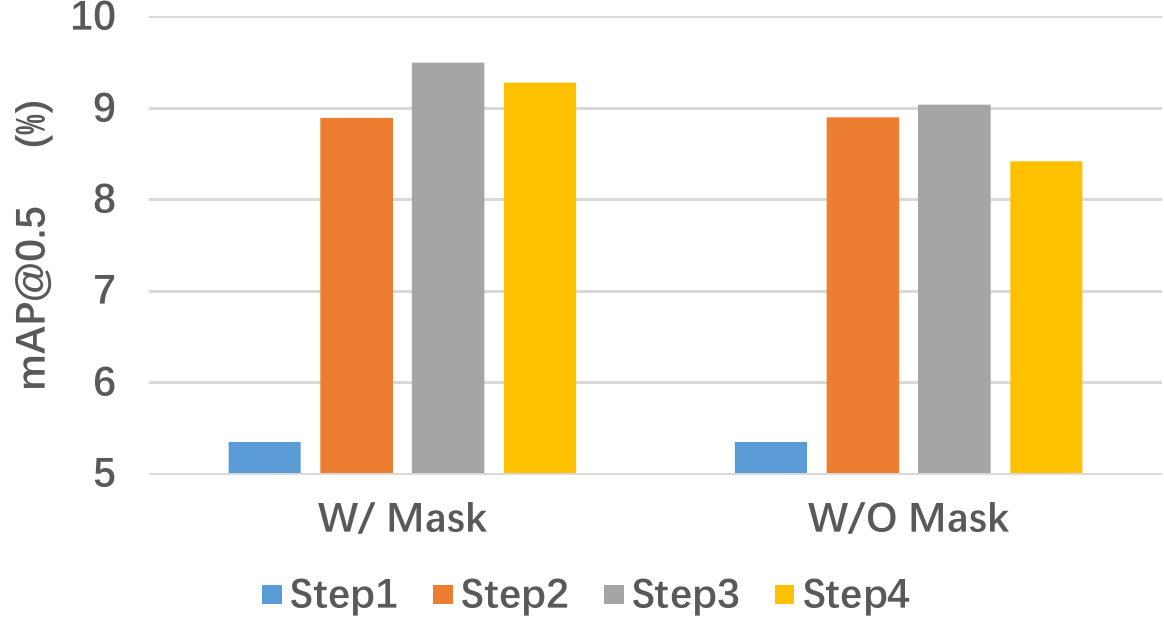} 
} 
\caption{\textit{Ablation about soft mask and erasion steps.}}
\label{fig:ablation1}
\end{figure}

\textbf{Ablation about soft mask and erasion steps.}
Firstly, we evaluate the utility of \emph{step-by-step} erasion. After a certain number of erasing steps, we directly take average of predictions as Eq.~\ref{eq6} and Eq.~\ref{eq7} on test data for evaluation. As shown in the \textit{left-hand side} of Figure~\ref{fig:ablation1}, a series of erasing operations indeed improves the detection performance from 5.3\% to 9.5\% mAP@0.5 on \textit{ActivityNet}, and from 6.9\% to 9.9\% mAP@0.5 on \textit{THUMOS'14}. However, \textit{excessive erasion} may introduce many false positive predictions and reduce the precision, so the mAP@0.5 declines after the \(4^{th}\) step on both datasets. To investigate the necessity of soft masks, we also report results without the mask on the \textit{right-hand side} of Figure~\ref{fig:ablation1}. From the side-by-side comparisons in the figure, we observe that the soft mask plays a role in two aspects. For one thing, as mentioned in the subsection~\ref{train_text}, it can suppress the detection score of background clips, so it mitigates the performance degradation from \textit{excessive erasion}. For another, it also imposes a tougher standard to select erasing snippets, and thus the results with mask at the early steps are slightly inferior to those without mask. Seeing as the tougher standard favors discriminative predictions with more certainty, this reverses at later steps, and the performance with mask is better than those without mask at last. 

\begin{figure*}[!t]\centering
\subfigure[\textit{mAP@0.5 at different steps.}] {
\includegraphics[width=0.85\textwidth]{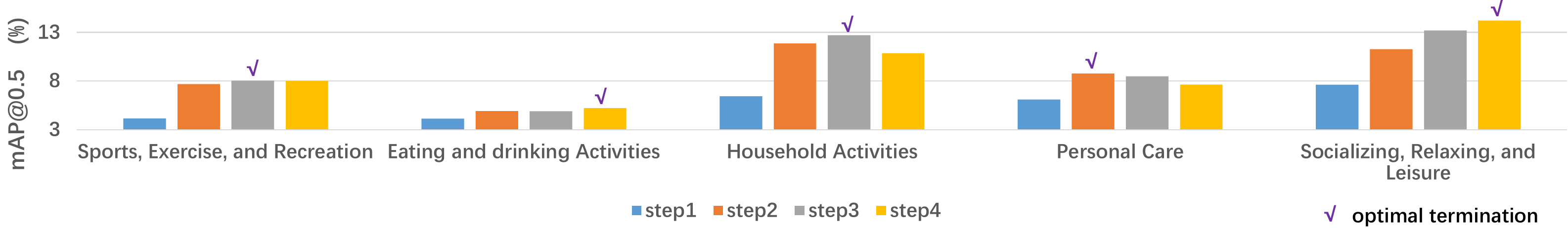}\label{stop_a}
} 
\subfigure[\textit{The value of \({m_j^T}\) at different steps.}] {
\includegraphics[width=0.85\textwidth]{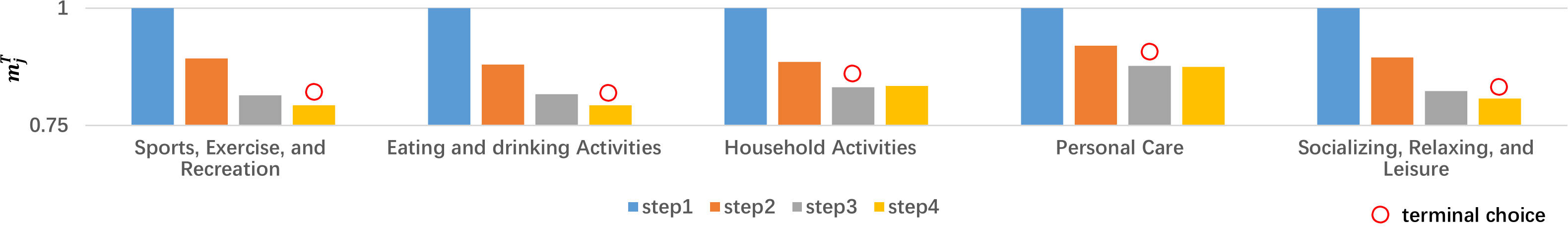}\label{stop_b}
} 
\caption{\textit{Discussion on training termination.} We report the \({m_j^T}\) in groups of the top-level category on \textit{ActivityNet}, and evaluate the detection performance by mAP@0.5.}\label{stop_condition}
\end{figure*}

\textbf{Discussion on training termination.}
As mentioned above, \textit{excessive erasion} has a negative influence on detection performance. To this end, it is of significance to find a criterion for erasing termination. In subsection~\ref{train_text}, we propose a criterion \({m_j^T}\) as Eq.~\ref{eq5} for the \(j^{th}\) category at the \(T^{th}\) step. We evaluate its effectiveness on \textit{ActivityNet}, and report the \({m_j^T}\) over the \(5\) categories of its top-level hierarchy for an intiutive illustration. For each top-level category, the value of mAP@0.5 and \({m_j^T}\) are calculated using the average of its subclasses. As Figure~\ref{stop_condition} depicts, the obvious degradation of detection performance occurs with the nearly invariable \({m_j^T}\), and we terminate erasing as shown in Figure~\ref{stop_b}. In this case, 3 out of 5 classes are ceased to be trained at the optimal step as Figure~\ref{stop_a} depicts. The other 2 classes achieve a close second-best performance, in which the mAP@0.5 is inferior to that of the best by less than \(0.4\%\). Since only given video-level category labels, we cannot always stop at the optimal step for every class. The criterion \({m_j^T}\) is simple yet effective to some extent, and at least prevents detection performance from suffering heavy loss. In the future, we may try on a more advanced terminal criterion.

\begin{figure}[!t]
\centering
\subfigure[\textit{THUMOS'14}.] {
\includegraphics[width=0.45\textwidth]{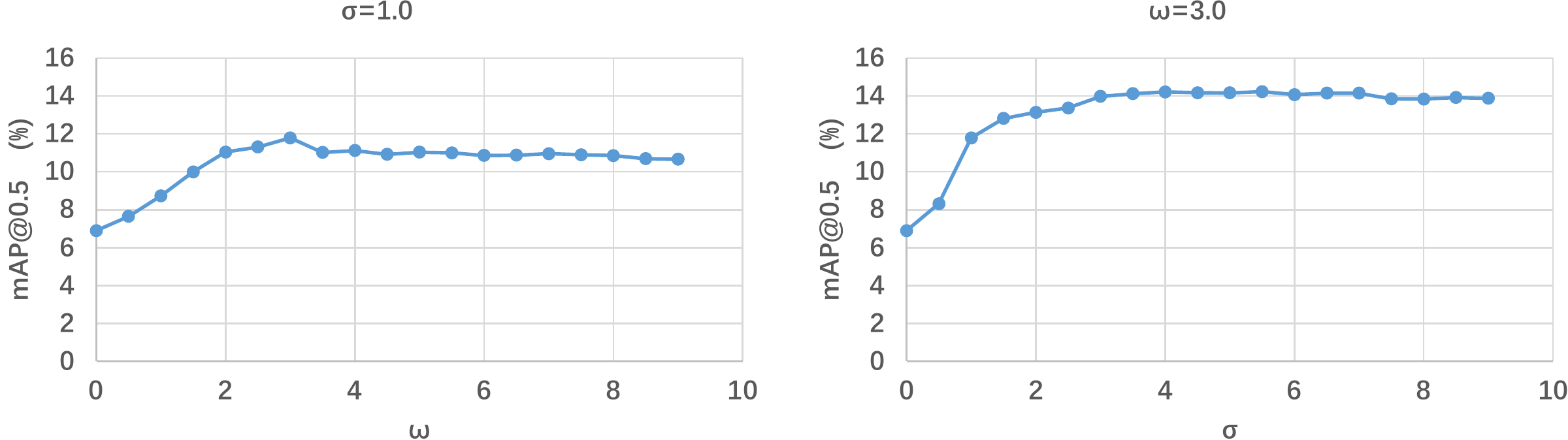} 
} 
\subfigure[\textit{ActivityNet}.] {
\includegraphics[width=0.45\textwidth]{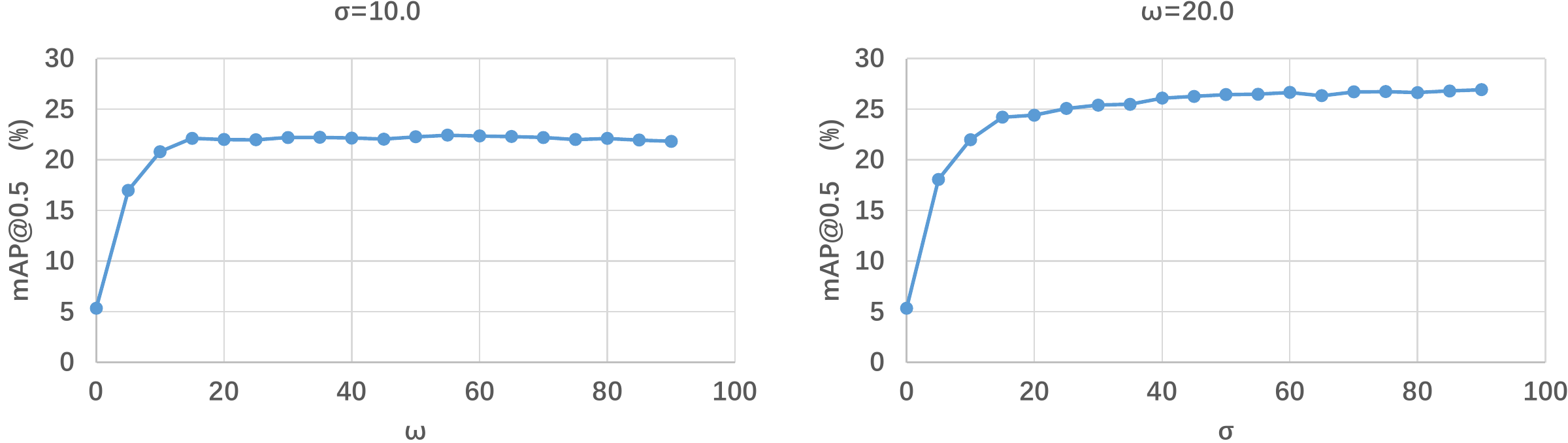} 
} 
\caption{\textit{Hyper-parametric stability of FC-CRF.} On both datasets, we evaluate the performance by mAP@0.5. The models on the left are with different \(\omega\) and a fixed \(\sigma\), while the models on the right are with different \(\sigma\) and a fixed \(\omega\).}
\label{fig:crf_param}
\end{figure}

\textbf{Effectiveness of FC-CRF and its hyper-parametric stability.} In the test phase, there are two crucial hyper-parameters w.r.t. our FC-CRF: \(\omega\) dominates the weight of pairwise potential and \(\sigma\) handles the scale of Gaussian kernel. Both are essential to the FC-CRF. Thus, we carry out two experiments at the first training step to evaluate the sensitivity of the two hyper-parameters on each dataset. On \textit{THUMOS'14}, we first fix \(\sigma\) to \(1.0\) and vary \(\omega\) from \(0\) to \(9.0\). As \textit{ActivityNet} has different sampling strategies and soft mask settings, we choose different hyper-parametric ranges in the first experiment: \(\sigma=10.0\) and \(\omega \in [0, 90.0]\). The results are shown in the \emph{left-hand part} of Figure~\ref{fig:crf_param}. It is quite evident that simply fusing the detection scores (in this case \(\omega=0\)) is not an appropriate choice, leading to a poor mAP performance. By properly choosing the value of \(\omega\), we can significantly improve the detection performance, and the performance remains highly stable across a wide range of \(\omega\). In the second the experiment, we fix the setting of \(\omega\) and change the value of \(\sigma\). We fix \(\omega=3.0\) on \textit{THUMOS'14} and \(\omega=20.0\) on \textit{ActivityNet}. As illustrated in the \emph{right-hand part} of Figure~\ref{fig:crf_param}, a proper \(\sigma\) can remarkably boost the detection performance. Likewise, the performance is highly stable across a wide range of \(\sigma\). To quantitatively demonstrate the effectiveness of our FC-CRF, we also report the mAP at various tIoU thresholds for a pair of suitable hyper-parameters in Table~\ref{crf_effect}. The FC-CRF drastically increases mAP@0.5 by \(20.8\%\) and \(7.1\%\) on \textit{ActivityNet} and \textit{THUMOS'14} respectively.

\begin{table}
\caption{{Effectiveness of FC-CRF.}}\label{crf_effect}  
\centering  
\subtable[mAP@tIoU on \textit{THUMOS'14.} \(\omega=3.0, \sigma=3.0.\)]{  
       \begin{tabular}{cccccc p{2cm}}
     \hline
     tIoU              & 0.5 & 0.4 & 0.3 & 0.2 & 0.1 \\ 
     \hline
     W/{\qquad}FC-CRF  & 14.0 & 20.4 & 28.5 & 36.3 & 42.9 \\ 
     W/O{\quad} FC-CRF  & 6.9 & 11.9 & 19.3 & 28.2 & 37.8 \\ 
     \hline
   \end{tabular}
}  
\qquad  
\subtable[mAP@tIoU on \textit{ActivityNet.} \(\omega=20.0, \sigma=40.0.\)]{       
  \begin{tabular}{ccccc}
    \hline
    tIoU              & Avg. & 0.95 & 0.75 & 0.5\\ 
    \hline
    W/{\qquad}FC-CRF  & 14.9 & 2.6 & 14.1 & 26.1 \\ 
    W/O{\quad} FC-CRF & 2.6 & 0.38 & 2.1 & 5.3 \\ 
    \hline
  \end{tabular}
}  
\end{table} 

\subsection{Evaluation of Temporal Action Detection}
In this subsection, we focus on the temporal action detection performance of our weakly supervised model.

\begin{figure*}[!t]
\centering\includegraphics[width=0.85\textwidth]{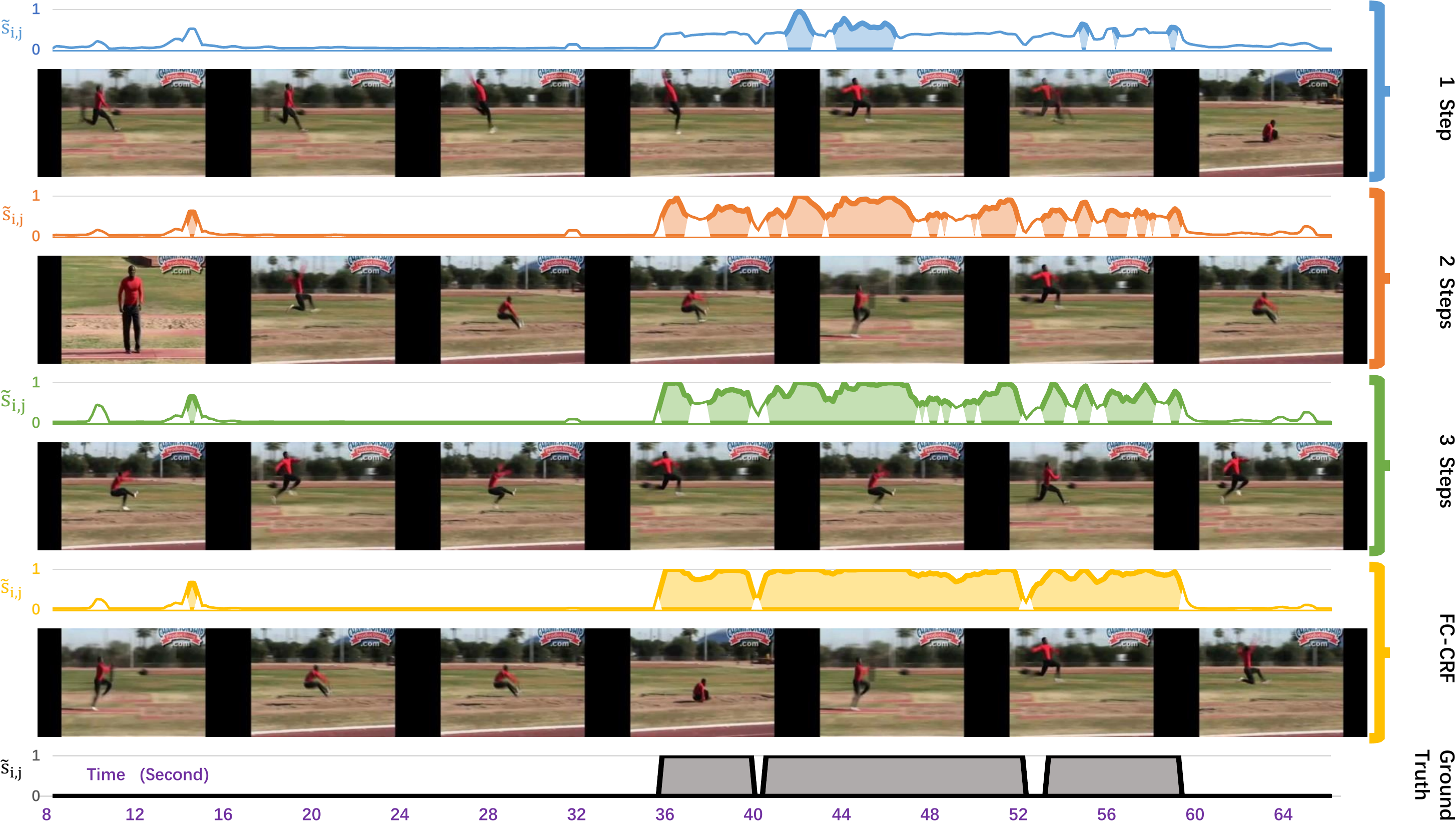}
\caption{\textit{Visualization of the detection process.} The snippet is excerpted from ``video\_test\_0001281'' between 8 and 67 seconds on \textit{THUMOS'14}. After a certain number of erasing steps, the curve of detection confidence \(\widetilde{s}_{i,j}\) is plotted for the action ``LongJump''. Under the confidence curve, a series of video frames discovered at the current step is on exhibition. The shaded areas underneath these curves represents the detected video clips (with \(\widetilde{s}_{i,j} > 0.5\)) up to the given erasing steps.}\label{final_test}
\end{figure*}

\textbf{Qualitative results.} We first visualize the learning process of our detector in Figure~\ref{final_test}. We can observe that a series of erasing steps facilitates the process to generate an integral video segment with action instance. Then the FC-CRF \textbf{\emph{retrieves the missed prediction}} occurring within the ground-truth segments, and \textbf{\emph{moderates noises caused by background snippets}} occurring within about 10-11, 31-32 and 65-66 seconds to some extent. It is worth mentioning that there is an interesting failure case approximately from 14 to 16 seconds in the video. In these two seconds, a coach demonstrated the run-up technique, but did not actually complete the whole long-jump activity. As human beings, we can easily distinguish this from the real long-jump. However, the detector mistakes this snippet possibly because it is difficult for classification networks to reason the temporally contextual relationship. In the area of video understanding, researchers still have a long way to go to enhance such a reasoning ability for recognition models.

\textbf{Quantitative results.} We finally report the performance of our detector, and make comparisons with state-of-the-art weakly supervised methods as well as strongly supervised approaches. For a temporal action detection task, \textbf{\textit{weak supervision}} refers to the setting for which only video-level category labels are provided, while \textbf{\textit{strong supervision}} refers to that both instance-level action categories and temporal boundary annotations are available. The results on the two datasets are shown in Table~\ref{thumos_det} and Table~\ref{anet_det}. The performance of our detector is superior to other weakly supervised methods. Compared with strongly supervised approaches, our model still achieves competitive performance, and even outperforms several of them.
\begin{table}[h] 
\centering
\caption{mAP@tIoU on \textit{THUMOS'14}.}\label{thumos_det}
\begin{threeparttable}
 \begin{tabular}{rccccc p{2cm}}
     \hline
     tIoU                    & 0.5 & 0.4 & 0.3 & 0.2 & 0.1 \\ 
     \hline
     \multicolumn{6}{l}{\textbf{Strong Supervision}} \\
     Karaman \textit{et al.}~\cite{KaramanFast} & 0.9 & 1.4 & 2.1 & 3.4 & 4.6 \\
     Wang \textit{et al.}~\cite{WangAction}    & 8.5 & 12.1& 14.6& 17.8& 19.2\\
     Heilbron \textit{et al.}~\cite{Heilbron2016Fast}& 13.5 & 15.2& 25.7& 32.9& 36.1\\
     Escorcia \textit{et al.}~\cite{Escorcia2016DAPs}& 13.9 &------&------&------&------\\
     Oneata \textit{et al.}~\cite{Dan2014The}  & 14.4& 20.8& 27.0& 33.6& 36.6\\
     Richard \textit{et al.}~\cite{Richard2016Temporal} & 15.2& 23.2& 30.0& 35.7& 39.7\\
     Yeung \textit{et al.}~\cite{yeung2016end}   & 17.1& 26.4& 36.0& 44.0& 48.9\\
     Yuan \textit{et al.}~\cite{Yuan2017Temporal}    & 17.8& 27.8& 36.5& 45.2& 51.0\\
     Yuan \textit{et al.}~\cite{Yuan2016Temporal}    & 18.8& 26.1& 33.6& 42.6& 51.4\\
     Shou \textit{et al.}~\cite{shou2016temporal}    & 19.0& 28.7& 36.3& 43.5& 47.7\\
     Shou \textit{et al.}~\cite{Shou2017CDC}    & 23.3& 29.4& 40.1& ------ & ------\\
     Lin \textit{et al.}~\cite{Lin2017Single}     & 24.6& 35.0& 43.0& 47.8& 50.1\\
     Xiong \textit{et al.}~\cite{xiong2017pursuit}\tnote{I}& 28.2& 39.8& 48.7& 57.7& 64.1\\
     Zhao \textit{et al.}~\cite{Zhao2017Temporal}\tnote{II}& 29.1& 40.8& 50.6& 56.2& 60.3\\    
     \hline\hline
     \multicolumn{6}{l}{\textbf{Weak Supervision}} \\
     Sun \textit{et al.}~\cite{Sun2015Temporal}  & 4.4 & 5.2 & 8.5 & 11.0 & 12.4\\

     Wang \textit{et al.}~\cite{wang2018untrimmed}         & 13.7 & 21.1 & 28.2 & 37.7 & 44.4 \\
     Ours                    & 15.9 & 22.5 & 31.1 & 39.0 & 45.8 \\  
     \hline
     
   \end{tabular}
   \begin{tablenotes}
        \footnotesize
        \item[I] {They use an actionness classifier trained on \textit{ActivityNet} for proposal generation.}
        \item[II] {They filter the detection results with the \textit{UntrimmedNets} to keep only those from the top-2 predicted action classes.}
   \end{tablenotes}
\end{threeparttable}
\end{table}
   
\begin{table}
  \caption{mAP@tIoU on \textit{ActivityNet}.}\label{anet_det}
  \begin{tabular}{rlcccc}
    \hline
     tIoU      &              & Avg. & 0.95 & 0.75 & 0.5\\ 
    \hline
    \multicolumn{6}{l}{\textbf{Strong Supervision}} \\
    \multirow{2}{*}{Xiong \textit{et al.}~\cite{xiong2017pursuit}}
    &One Stage&------&------&------&   9.0\\
    &Cascade  &24.9  &   5.0&  24.1&  41.1\\
    \multirow{2}{*}{Zhao \textit{et al.}~\cite{Zhao2017Temporal}} 
    &SW-SSN   &18.2&------&------&------\\
    &TAG-SSN  &24.5&------&------&------\\
    \hline\hline
    \multicolumn{6}{l}{\textbf{Weak Supervision}} \\
    Ours&     &15.6&   2.9&  14.7&   27.3\\ 
    \hline
  \end{tabular}
\end{table}

\section{Conclusion}
In this paper, we address the problem of weakly supervised temporal action detection in untrimmed videos. Given only video-level category labels, we utilize a series of classifiers to detect discriminative temporal regions. Specifically, the series of classifiers are built with \emph{step-by-step} erasion on snippets with high detection confidence from the remaining video data. In test process, we expediently collect predictions from the \emph{one-by-one} classifiers. Moreover, we introduce an FC-CRF for imparting prior knowledge to our detector. Notwithstanding the prior knowledge is simply based upon temporal coherence, the FC-CRF significantly improves the detection performance. Extensive experiments on two challenging datasets illustrate that our approach achieves superior performance to state-of-the-art weakly supervised results, and is also comparable to many strongly supervised methods.


\newpage

\bibliographystyle{ACM-Reference-Format}
\bibliography{sample-bibliography}

\end{document}